\documentclass[10pt,twocolumn,letterpaper]{article}

\usepackage{iccv} % To produce the CAMERA-READY version
\definecolor{iccvblue}{rgb}{0.21,0.49,0.74}
\usepackage[pagebackref,breaklinks,colorlinks,allcolors=iccvblue]{hyperref}

\def\paperID{}
\def\confName{ICCV}
\def\confYear{2025}

\title{BayesSDF: Surface-Based Laplacian Uncertainty Estimation for 3D Geometry with Neural Signed Distance Fields}

\author{
Rushil Desai\\
Berkeley Artificial Intelligence Research\\
University of California, Berkeley\\
{\tt\small rushildesai@berkeley.edu}
}

\begin{document}
\maketitle
\begin{abstract}
Accurate surface estimation is critical for downstream tasks in scientific simulation, and quantifying uncertainty in implicit neural 3D representations still remains a substantial challenge due to computational inefficiencies, scalability issues, and geometric inconsistencies. However, current neural implicit surface models do not offer a principled way to quantify uncertainty, limiting their reliability in real-world applications. Inspired by recent probabilistic rendering approaches, we introduce BayesSDF, a novel, probabilistic framework for uncertainty estimation in neural implicit 3D representations. Unlike radiance-based models such as Neural Radiance Fields (NeRF) or 3D Gaussian Splatting, Signed Distance Functions (SDFs) provide continuous, differentiable surface representations, making them especially well-suited for uncertainty-aware modeling. BayesSDF applies a Laplace approximation over SDF weights and derives Hessian-based metrics to estimate local geometric instability. We empirically demonstrate that these uncertainty estimates correlate strongly with surface reconstruction error across both synthetic and real-world benchmarks. By enabling surface-aware uncertainty quantification, BayesSDF lays the groundwork for more robust, interpretable, and actionable 3D perception systems.
\end{abstract}
    
\section{Introduction}
\label{sec:intro}

Scientific applications such as fluid simulation, medical reconstruction, and robotic perception require not only accurate surface reconstructions but also reliable uncertainty estimates \cite{abdar2021review} to support robust decision-making (e.g. aleatoric vs epistemic \cite{kendall2017uncertainties}). Recent advances in multi-view 3D reconstruction using neural implicit representations \cite{mildenhall2020nerf} have dramatically improved surface quality, yet these methods are rarely adopted in the real world \cite{martinbrualla2021nerfw} due to their inability to model surface uncertainty. Neural implicit representations model geometry as continuous functions \cite{mildenhall2020nerf, barron2021mipnerf, barron2022mipnerf360}, enabling high-resolution reconstructions without relying on dense voxels or explicit meshes. While these methods, like Neural Radiance Fields \cite{mildenhall2020nerf}, are effective at view synthesis, they lack surface-level structure and provide no mechanism for quantifying geometric uncertainty. Models like NeuS \cite{wang2021neus} (and dynamic topological models like Nerfies \cite{park2021nerfies}, HyperNeRF \cite{park2022hypernerf}, CamP \cite{park2022campnerf}, etc.) address this limitation by using Signed Distance Functions (SDFs), which define surfaces as zero-level sets of continuous scalar fields. This provides strong geometric structure and differentiability, enabling more consistent and physically meaningful reconstructions. However, these models remain deterministic and offer no mechanism for estimating uncertainty over the learned surface SDF fields. To fill this gap, we introduce BayesSDF, a probabilistic framework for estimating surface-aligned uncertainty in neural implicit models that use SDFs. Our work is inspired by BayesRays \cite{goli2024bayesrays}, which introduced uncertainty modeling for volumetric radiance fields via Bayesian rendering. However, BayesSDF departs fundamentally from BayesRays by focusing on explicit surface geometry. Instead of modeling uncertainty through pixel-space or volumetric sampling, BayesSDF leverages a Laplace approximation \cite{daxberger2021laplaceredux} over SDF weights to quantify local surface sensitivity using second-order derivatives. This results in uncertainty maps that are geometrically coherent and directly tied to surface stability. By combining the structure of SDF-based models with a tractable probabilistic approximation, BayesSDF produces uncertainty estimates that strongly correlate with reconstruction error, offering more interpretable and actionable outputs. These estimates help identify ambiguous or ill-constrained geometry, which is critical for downstream tasks that require reliable spatial reasoning \cite{wilson2023visualizingsba}. Unlike uncertainty models based on pixel-space rendering or volumetric sampling, BayesSDF provides surface-aware estimates rooted in the geometry itself. This enables more efficient and scalable modeling without sacrificing spatial precision.
To summarize, our key contributions are as follows:
\begin{itemize}
\item \textbf{Surface-Aware Uncertainty via SDF Geometry}: Prior methods often model uncertainty indirectly in pixel or depth space, which can misalign with true geometric errors. BayesSDF instead leverages the analytic gradients of Signed Distance Functions to estimate uncertainty directly on the surface. This direct integration enables uncertainty estimates that are physically meaningful and geometrically consistent.
\item \textbf{Uncertainty Estimation through Local Surface Sensitivity}: BayesSDF uses a Laplace approximation and Hessian-based calculations over the neural SDF weights, capturing how local geometric perturbations affect surface stability in the geometry's deformation field. By evaluating how small perturbations in geometric parameters affect rendered color outputs, the method accurately quantifies spatial uncertainty, avoiding the computational expense of traditional ensemble \cite{lakshminarayanan2017ensemble} or extensive sampling approaches.
\item \textbf{Scalable Deformation Modeling with Hash Encoding}: To model local geometric variation, BayesSDF introduces a hierarchical multi-resolution hash-encoded deformation field. Compared to grid-based representations used in prior probabilistic methods like BayesRays \cite{goli2024bayesrays}, our approach supports higher spatial resolution with significantly reduced memory overhead. This design makes BayesSDF scalable to complex scenes while remaining compatible with frameworks like NeuS \cite{wang2021neus}.
\end{itemize}
Motivated by computational physics and scientific simulation applications with 3D environments such as modeling fluid flow through forests, where precise surface geometry and reliable uncertainty estimates are essential. SDF-based models, as demonstrated in DeepSDF \cite{park2019deepsdf}, provide surface continuity and differentiability that are advantageous for physical modeling. The proposed method demonstrates superior performance across multiple synthetic and real-world datasets, achieving state-of-the-art uncertainty calibration and reliability in 3D surface modeling. Our framework explicitly supports such use cases by relating uncertainty to local surface instability.

\section{Related Work}
\label{sec:relatedwork}
Uncertainty estimation is essential in 3D learning systems, especially for safety-critical and scientific applications, where downstream tasks depend on the reliability of reconstructed geometry. Traditional neural implicit methods such as Neural Radiance Fields (NeRF) \cite{mildenhall2020nerf, martinbrualla2021nerfw} and Signed Distance Function (SDF)-based models like NeuS \cite{wang2021neus} have achieved high-fidelity reconstruction \cite{zhang2020nerfplusplus, tancik2020fourier} and view synthesis, but largely ignore uncertainty modeling. This omission limits their usefulness in tasks requiring interpretable or risk-aware decision-making.
In this section, we contextualize BayesSDF, within three major threads of prior work:
\begin{enumerate}
    \item Why SDF-based models \cite{park2019deepsdf, wang2021neus} offer superior structure for geometric uncertainty estimation compared to NeRF-style density fields. \cite{mildenhall2020nerf, martinbrualla2021nerfw, chen2021mvsnerf}
    \item How uncertainty has been previously modeled in neural 3D systems, including ensemble methods and Bayesian inference.
    \item Why BayesSDF provides a principled and non-trivial improvement over prior Bayesian extensions like BayesRays \cite{goli2024bayesrays}, particularly in aligning uncertainty with surface structure.
\end{enumerate}
\subsection{Uncertainty Estimation in 3D Reconstruction Methods}
3D reconstruction methods vary widely in how they represent geometry, which directly impacts how uncertainty can be quantified. NeRF-style models \cite{mildenhall2020nerf, barron2021mipnerf, barron2022mipnerf360} rely on volume densities sampled along camera rays. These are well-suited for view synthesis but decouple uncertainty from physical surfaces. In contrast, SDF-based models \cite{park2019deepsdf, smith2013uq, seyb2019nonlinearspheretracing, abou2022freenerf, boss2021nerv, jiang2022gnt} define surfaces as the zero level-set of continuous scalar fields, offering geometric interpretability and a natural differentiable structure \cite{oechsle2021unisurf, rematas2022urban, yariv2021volsdf}, which are features critical for scientific computing tasks involving simulations, collisions, or boundary conditions.
BayesSDF builds on this structure by estimating uncertainty not in ray-space but in the SDF’s parameter space. This enables surface-aware uncertainty aligned with real geometric features rather than visual artifacts.
\subsubsection{Neural Representations and SDF-Based Methods}
Neural implicit representations, such as NeRF \cite{mildenhall2020nerf} and NeuS \cite{wang2021neus}, have demonstrated state-of-the-art performance in view synthesis and geometric reconstruction. However, their uncertainty estimation techniques differ significantly.
\begin{itemize}
\item NeRF-based methods \cite{mildenhall2020nerf, shen2021snerf, shen2022cfnerf} estimate the uncertainty per ray by analyzing the variance in color and density outputs across multiple samples along a viewing ray. These approaches are effective for rendering applications, but they do not capture geometric uncertainty directly \cite{bae2021aleatoricuncertainty}.
\item NeuS-based methods \cite{wang2021neus} instead represent surfaces using an SDF, which provides a continuous and differentiable description of 3D geometry. \cite{sucar2021imap, xu2022geonerf} However, existing NeRF \cite{mildenhall2020nerf} uncertainty estimation techniques do not translate well to NeuS \cite{wang2021neus} because they rely on ray-space density variations rather than surface-level geometric uncertainty.
\end{itemize}
\subsection{Ensemble-Based Methods and Their Limitations}
Ensemble-based methods \cite{lakshminarayanan2017ensemble, abdar2021review, kendall2017uncertainties} are a standard approach to approximate epistemic uncertainty by training and sampling multiple networks. In the context of reconstruction methods, ensemble methods are particularly advantageous as they offer robust uncertainty quantification without requiring complex probabilistic modeling or extensive sampling. While ensemble methods are widely used in 2D vision tasks (e.g. image classification and segmentation), they introduce several limitations in the 3D domain:
\begin{itemize}
\item \textbf{Scalability:} Training multiple neural implicit models is memory and compute intensive, particularly for high-resolution 3D reconstructions. \cite{mueller2022instantngp, chen2022mobilenerf, yu2021plenoctrees, tancik2023nerfstudio} Traditional ensemble methods require storing and evaluating multiple models, leading to scalability issues \cite{warburg2023nerfbusters}.
\item \textbf{Surface Consistency:} Unlike 2D image-based tasks \cite{monteiro2020ssn}, uncertainty in 3D learning must be estimated consistently across a continuous surface rather than at discrete pixel locations. \cite{zhu2022nice, zhan2022activermap}
\item \textbf{Ray vs. Volume-Based Estimation:} Many existing ensemble-based uncertainty estimation techniques in NeRF-like \cite{mildenhall2020nerf} models operate strictly along rays. This results in uncertainty that does not align well with geometric structures, making it less interpretable.
\end{itemize}
These challenges motivate more principled Bayesian approximations within neural 3D representations leveraging the strengths of ensemble-based learning, particularly those offering geometric continuity like SDFs \cite{yan2022activeneuralmapping}.
\subsection{Bayesian Inference for Ray-Based Sampling}
BayesRays \cite{goli2024bayesrays} introduced uncertainty estimation in NeRF \cite{mildenhall2020nerf} by modeling pixel-space color and density distributions using a Bayesian ensemble and ray sparsification. While BayesRays \cite{goli2024bayesrays} is effective for capturing radiance uncertainty, it remains tied to volumetric density representations and does not explicitly model surface uncertainty \cite{shen2021snerf, pumarola2020dnerf, cao2022tnerf, zhang2023mvsnerfplusplus, riegler2023nerftex, hoffman2023probnerf}. It relies on ray-space sparsification techniques \cite{park2021nerfies, park2022hypernerf, newcombe2015dynamicfusion, tretschk2021nrnerf}, which can lead to misalignment between estimated uncertainty and actual geometric uncertainty \cite{tagliasacchi2023volumerenderingdigest}. Its volumetric density representation lacks an explicit notion of surface, making it harder to isolate geometric ambiguity.
BayesSDF addresses these issues by modeling the sensitivity of the surface geometry itself. Rather than relying on variation in rendered pixel values, BayesSDF quantifies how local perturbations in the SDF field affect surface stability using a Laplace approximation \cite{daxberger2021laplaceredux, smith2013uq, seyb2019nonlinearspheretracing, niemeyer2020differentiable, xu2022learning, ritter2018scalablelaplace}. This results in uncertainty maps that are tightly coupled to the surface \cite{park2019deepsdf, neal1995bayesian}, making them more interpretable and physically meaningful. Notably, directly applying BayesRays \cite{goli2024bayesrays} to SDF-based models like NeuS \cite{wang2021neus} is non-trivial. BayesRays \cite{goli2024bayesrays} relies on sampling distributions along rays \cite{park2019deepsdf, niemeyer2022regnerf}, which do not generalize to SDF’s zero-crossing surface logic. In contrast, BayesSDF incorporates the analytic structure of SDFs via second-order derivatives \cite{daxberger2021laplaceredux, smith2013uq, neal1995bayesian}, enabling uncertainty estimates that are both tractable and geometrically consistent.

\begin{table}[ht]
    \centering
    \renewcommand{\arraystretch}{1.2}
    \resizebox{\columnwidth}{!}{
    \begin{tabular}{lcccc}
        \toprule
        \textbf{Method} & \textbf{Representation} & \textbf{Uncertainty Estimation} & \textbf{Computational Cost} & \textbf{Geometric Consistency} \\
        \midrule
        \textbf{NeRF \cite{mildenhall2020nerf}} & Volume Density & None & High & Low \\
        \textbf{NeuS \cite{wang2021neus}} & SDF & None & High & High \\
        \textbf{BayesRays \cite{goli2024bayesrays}} & Volume Density & Ray-Space Variance & Moderate & Moderate \\
        \textbf{BayesSDF} & SDF & Hessian-Based Laplace Approx. & Low & Very High \\
        \bottomrule
    \end{tabular}
    }
    \caption{Comparison of existing 3D reconstruction methods based on geometry representation, uncertainty modeling, computational cost, and surface-level interpretability.}
    \label{tab:comparison}
\end{table}

\section{Background}
\label{sec:background}
Quantifying uncertainty in 3D implicit representations is essential to improve the reliability and interpretability of learned geometric models. This section reviews the foundational components upon which BayesSDF builds upon.
\subsection{Neural Implicit Surfaces (NeuS)}
Signed Distance Functions (SDFs) define geometry as a continuous field that measures the signed distance of a point in space to the closest surface boundary. Formally, an SDF is defined as \( f: \mathbb{R}^3 \to \mathbb{R} \), where:
\[
f(\mathbf{x}) = \begin{cases}
\phantom{-}d, & \text{if } \mathbf{x} \text{ is outside the surface} \\
\phantom{-}0, & \text{if } \mathbf{x} \text{ lies on the surface} \\
-d, & \text{if } \mathbf{x} \text{ is inside the surface}
\end{cases}
\]
This representation allows for accurate computation of surface normals and curvature, and forms the basis of many modern 3D reconstruction methods \cite{park2019deepsdf, boss2021nerd, jiang2022gnt, rematas2022urban, oechsle2021unisurf, yariv2021volsdf, niemeyer2020differentiable}. Despite their geometric advantages, SDFs have not been fully explored for probabilistic modeling or uncertainty estimation in learned 3D representations \cite{lee2024bayesiannerf, hoffman2023probnerf, yan2022activeneuralmapping}.
NeuS \cite{wang2021neus} addresses the geometric limitations of NeRF \cite{mildenhall2020nerf} by replacing volumetric density fields with SDFs that model surfaces as zero-level sets. The rendering is adapted by transforming SDF values into opacity via a sigmoid function:
\[
\alpha(\mathbf{x}) = \Phi_s(f(\mathbf{x})) = \frac{1}{1 + e^{-s f(\mathbf{x})}}
\]
The final color is computed using transmittance-weighted rendering:
\[
C(\mathbf{r}) = \sum_{i} T_i \alpha_i \mathbf{c}_i, \quad T_i = \prod_{j=1}^{i-1} (1 - \alpha_j)
\]
NeuS \cite{wang2021neus} enables precise geometric modeling, but lacks inherent mechanisms for uncertainty quantification.
\subsection{Uncertainty Estimation via Ensembles}
Ensemble learning estimates epistemic uncertainty by aggregating predictions from multiple independently trained models. For a set of \( N \) models, the mean and variance of predictions at point \( \mathbf{x} \) are:
\[
\bar{f}(\mathbf{x}) = \frac{1}{N} \sum_{i=1}^{N} f_i(\mathbf{x}), \quad
\text{Var}(\mathbf{x}) = \frac{1}{N} \sum_{i=1}^{N} (f_i(\mathbf{x}) - \bar{f}(\mathbf{x}))^2
\]
This variance captures model disagreement, which is indicative of uncertainty \cite{lakshminarayanan2017ensemble, kendall2017uncertainties, bae2021aleatoricuncertainty, zaidi2022neuralensemble}. While ensemble methods are effective across 2D and 3D vision tasks \cite{abdar2021review}, they are resource-intensive and can be misaligned with underlying geometry in volumetric or ray-based models \cite{zhu2022nice, yu2022mononerd}.
% \subsection{Bayesian Uncertainty in NeRF: BayesRays}
% BayesRays proposes a Bayesian approach for NeRF by learning a density-aware variance model and propagating it along rays \cite{goli2024bayesrays}. The estimated uncertainty is given by:
% \[
% U(\mathbf{x}) = \frac{1}{K} \sum_{k=1}^{K} \sigma_k(\mathbf{x}) \cdot T_k(\mathbf{x})
% \]
% Color and density are modeled as Gaussian random variables, and uncertainty in the rendered image is computed by propagating these distributions through the rendering equation:
% \[
% \text{Var}[\hat{C}(\mathbf{r})] = \sum_{i=1}^N T_i^2 \left(1 - \exp(-\sigma_i \delta_i)\right)^2 \text{Var}[\mathbf{c}_i]
% \]
% BayesRays achieves efficient uncertainty estimation without ensembles, but its output lacks surface-level geometric consistency. Integrating such uncertainty with geometric fields remains a core challenge in the field \cite{kendall2017uncertainties}.

\section{Method}
\label{sec:method}
In this section, we detail our proposed BayesSDF method, which leverages a NeuS \cite{wang2021neus} scene representation and adopts an uncertainty estimation framework through a Hessian-based deformation field. We begin by describing the theoretical underpinnings, then present how we construct and query the deformation field, calculate the Hessian, and finally approximate local uncertainty. 
\subsection{Theory}
Our goal is to model an associated uncertainty measure derived from the local sensitivity of the rendering function to small perturbations in the underlying deformation parameters. Let \( \mathbf{x} \in \mathbb{R}^3 \) denote a 3D point in the scene space, and let \( \mathbf{c}(\mathbf{x}) \in \mathbb{R}^3 \) be the color (in red-green-blue (RGB)) rendered by the neural field. Rather than relying solely on a deterministic mapping \( \mathbf{x} \mapsto \mathbf{c}(\mathbf{x}) \), we attach a set of local deformation parameters \( \mathbf{d}(\mathbf{x}) \) that influence the geometry (and thus the color) through a learned distortion or deformation field \cite{pumarola2020dnerf, park2021nerfies}. Formally, we assume \(\mathbf{c}(\mathbf{x}) \;=\; \mathcal{R}\bigl(\mathbf{x}, \mathbf{d}(\mathbf{x}); \Theta\bigr)\) where \( \mathcal{R} \) is the rendering operator that depends on both the scene geometry (parameterized by a neural SDF model and any additional shape parameters \( \Theta) \) and the local deformation \( \mathbf{d}(\mathbf{x}) \). To estimate uncertainty, we measure how sensitively \( \mathbf{c}(\mathbf{x}) \) changes with respect to changes in \( \mathbf{d}(\mathbf{x}) \) \cite{kendall2017uncertainties, abdar2021review}. Intuitively, if small variations in \( \mathbf{d}(\mathbf{x}) \) induce large changes in color, we interpret that region as having higher uncertainty in the final scene reconstruction \cite{daxberger2021laplaceredux, ritter2018scalablelaplace}.
\subsection{Distortion Fields}
The distortion field \( f_{\mathrm{hash}}(\mathbf{x}) \) captures spatial distortions in the SDF field, serving as a basis for uncertainty estimation. To represent the deformation parameters \( \mathbf{d}(\mathbf{x}) \) in a continuous domain, we employ a hash encoding that functions as a voxel-like grid implemented through a hash table \(\mathbf{d}(\mathbf{x}) = f_{\mathrm{hash}}(\mathbf{x}) \in \mathbb{R}^3\) where \( f_{\mathrm{hash}} \) is a function learned by a single-level hash encoding with: \( \mathrm{min\_res} = \mathrm{max\_res} = 2^{\ell} \) for level of detail \( \ell \), a hash table size of \( 2^{3\ell + 1} \) entries, linear interpolation between grid corners, and 3 output channels per queried grid vertex \cite{mueller2022instantngp}. Effectively, for a given point \( \mathbf{x} \), the hash encoding looks up an 8-corner neighborhood around the voxel containing \( \mathbf{x} \), interpolates among these corners (each storing a 3D deformation vector) and outputs a single 3D vector \( \mathbf{d}(\mathbf{x}) \). 
% This approach is parameter-efficient while remaining flexible. We denote \( \mathbf{aabb} \in \mathbb{R}^6 \) as the scene bounding box, which is used to normalize \( \mathbf{x} \) to a unit cube or appropriate domain before querying the hash grid. Specifically, we generate a ray bundle from the dataset's rays, containing origins, directions, and near/far bounds. We then use Neus' hierarchical or “proposal” sampler to refine sample locations along each ray, focusing more samples near scene geometry (i.e., near the SDF zero level set). 
Let \( \{\mathbf{x}_i\} \) be the set of 3D points sampled on a ray. When each ray sample \( \mathbf{x} \in \mathbb{R}^3 \) is generated, we transform \( \mathbf{x} \) into normalized coordinates \( \tilde{\mathbf{x}} = \mathrm{normalize}(\mathbf{x}, \mathbf{aabb}) \). Then, \(\mathbf{d}(\mathbf{x})\;=\;\sum_{c=1}^{8} \alpha_c(\tilde{\mathbf{x}}) \,\mathbf{d}_c\quad\text{where}\quad\bigl\{\mathbf{d}c\bigr\}{c=1}^8\;\) are are the corner deformation vectors that are found from a grid cell index in each dimension (one for each binary combination of \( \lfloor\cdot\rfloor or \lceil\cdot\rceil \) in (x,y,z)) and \( \alpha_c(\tilde{\mathbf{x}}) \) are the trilinear interpolation coefficients \( (\alpha_c(\mathbf{x}i) \;\in\; [0,1], \quad c = 1,\ldots,8,) \) such that \( \sum{c=1}^8 \alpha_c(\mathbf{x}_i) = 1 \). These coefficients are used to blend the deformation vectors from the 8 corners. This procedure is vital not only for generating \( \mathbf{d}(\mathbf{x}_i) \), but also for accumulating partial derivatives. Because each point’s deformation influences eight corner entries, we track which corners are involved and in what proportion.
\subsection{Calculating the Hessian}
The uncertainty computation in BayesSDF revolves around accumulating partial derivatives of the rendered color \( \mathbf{c}(\mathbf{x}) = (r(\mathbf{x}),\,g(\mathbf{x}),\,b(\mathbf{x})) \) with respect to the deformation parameters stored in the hash grid. Because the deformation field is queried at multiple corners (8 corners per voxel), we track partial derivatives for each corner contribution. Let us denote the final, aggregated color at a particular sample by \(\mathbf{C} \;=\; \sum_{\mathbf{x} \in \mathcal{R}\text{-samples}} \mathbf{c}(\mathbf{x})\) where the sum is over all the sample points along the ray (or potentially aggregated over multiple rays). In the code, we backpropagate through each color channel independently to accumulate partial derivatives. For a single color channel \( c \in \{r,g,b\} \), we compute \(
\frac{\partial c}{\partial \mathbf{d}_c^{(k)}} \) where \( \mathbf{d}_c^{(k)} \) refers to the offsets (the deformation parameters) at corner \( k \in \{1,\dots,8\} \). We do this by invoking a backwards pass for each color channel and capturing the resulting gradients with respect to \( \mathbf{d} \). Each time, we zero out the gradient from the previous backward pass to isolate the new color channel's derivative. Between each step, the accumulated gradient is saved (and subsequently multiplied by interpolation coefficients, etc.), ensuring that the next backward pass only captures derivatives for the next channel. This pattern also addresses memory constraints: if we did not zero out or separate channel-wise backprop, the gradient buffers for the entire 3D field could become unwieldy. Once these partial derivatives are obtained, we aggregate them in a variable called `hessian` in the script. Concretely, for each corner’s index k in the hash grid, the squared sum of partial derivatives is computed as follows:
\[
\Delta(\mathbf{x})
\;=\;
\sum_{d=1}^{3}
\left(
\left(\frac{\partial r}{\partial \mathbf{d}_d}\right)^2 \cdot
\left(\frac{\partial g}{\partial \mathbf{d}_d}\right)^2 \cdot
\left(\frac{\partial b}{\partial \mathbf{d}_d}\right)^2
\right)
\]
where \( \mathbf{d}_d \in \{\mathbf{d}_1, \mathbf{d}_2, \mathbf{d}_3\} \) represents each dimension of the deformation vector. In short, we take the partial derivative of r, g, and b with respect to each dimension of the local offset, square them, and sum them to form a scalar representing local “curvature” in color space for that grid corner. We organize these per-corner accumulations into a 3D tensor of size \( ((2^\ell)+1)^3 \). Each cell in that 3D tensor corresponds to a location in the hash-encoded deformation grid. Over multiple training iterations, we repeatedly add these partial-derivative-based values, accumulating them into a global measure that, once normalized or post-processed, can be interpreted as the Hessian with respect to the neural color output and the deformation parameters.
\subsection{Distortion Space Perturbations}
A critical aspect of BayesSDF is its reliance on small local changes in the deformation field to gauge uncertainty. Formally, if \( \mathbf{d}(\mathbf{x}) \) is the deformation vector for the voxel corner associated with point \( \mathbf{x} \), we implicitly examine the infinitesimal perturbations \(\delta \mathbf{d}(\mathbf{x}) \;\in\; \mathbb{R}^3\) measure their effect on the rendered color  \( \mathbf{c}(\mathbf{x}) \). In other words, we approximate the local mapping
\[
\mathbf{c}\bigl(\mathbf{x};\,\mathbf{d}(\mathbf{x}) + \delta \mathbf{d}(\mathbf{x})\bigr)
\;\approx\;
\mathbf{c}\bigl(\mathbf{x};\, \mathbf{d}(\mathbf{x})\bigr)
\;+\;
\nabla_{\mathbf{d}(\mathbf{x})} \mathbf{c}\bigl(\mathbf{x}\bigr)
\,\delta \mathbf{d}(\mathbf{x}),
\]
where \( \nabla_{\mathbf{d}(\mathbf{x})}\mathbf{c}(\mathbf{x}) \) is precisely the gradient tensor we accumulate in the Hessian array. The magnitude of these partial derivatives quantifies the sensitivity of \( \mathbf{c}(\mathbf{x}) \) to local perturbations \( \delta \mathbf{d}(\mathbf{x}) \). In practice, perturbations are never explicitly applied to \( \mathbf{d}(\mathbf{x}) \); rather, their effect is simulated by the gradient computation (.backward()). This is central to the BayesSDF mechanism: we do not need to stochastically perturb each voxel corner. Instead, automatic differentiation yields \( \partial \mathbf{c}(\mathbf{x}) / \partial \mathbf{d}(\mathbf{x}) \) for each channel, which we then square and sum to approximate local uncertainty.
\subsection{Uncertainty Approximation}
We interpret these aggregated squared gradients in `hessian` as proxies for uncertainty. High squared partial derivatives indicate that small changes in the deformation field cause large changes to the rendered color \cite{lakshminarayanan2017ensemble, neal1995bayesian, lee2024bayesiannerf}, suggesting the reconstruction at that location is sensitive and potentially uncertain \cite{hoffman2023probnerf, goli2024bayesrays}. Conversely, near-zero partial derivatives indicate that changes in local deformation parameters barely affect the color, suggesting more stable or confident estimates. Given the final accumulated \( \mathbf{H} \) stored in our 3D grid, we define an uncertainty function \( \sigma \) over the domain:
\[
\sigma(\mathbf{x})
\;=\;
\sqrt{
\mathbf{H}\bigl(\text{grid\_index}(\mathbf{x})\bigr)
}
\]
Because \( \mathbf{H} \) is computed by adding the squared derivatives for the three color channels (each with respect to three offset dimensions), it encapsulates the total 'sensitivity' at a location. One may further process \( \sigma(\mathbf{x}) \) by normalization or thresholding to highlight regions of highest uncertainty.

\section{Results}
\label{sec:results}
In this section, we empirically evaluate BayesSDF and demonstrate its effectiveness as a scene reconstruction framework with principled uncertainty estimation. We describe our rendering pipeline, discuss the uncertainty evaluation metrics, and present quantitative and qualitative results, highlighting how BayesSDF is useful for calculating depth accuracy and uncertainty correlation in a geometric sense.
\subsection{Rendering Techniques}
We employ a custom rendering method that integrates seamlessly with BayesSDF to generate RGB, depth, and uncertainty images under various camera trajectories. This rendering process is a vital component of our experimental pipeline, enabling us to visualize and analyze uncertainty predictions alongside scene geometry. The method supports producing multiple outputs, most importantly: RGB images, Depth images (visualized with a colormap), and Uncertainty maps (also colormapped, typically “inferno”). Each pixel’s uncertainty value is fed through a user-specified color mapping function to produce interpretive 2D heatmaps of the 3D uncertainty field.
\begin{figure}[h]
    \centering
    \includegraphics[width=\columnwidth]{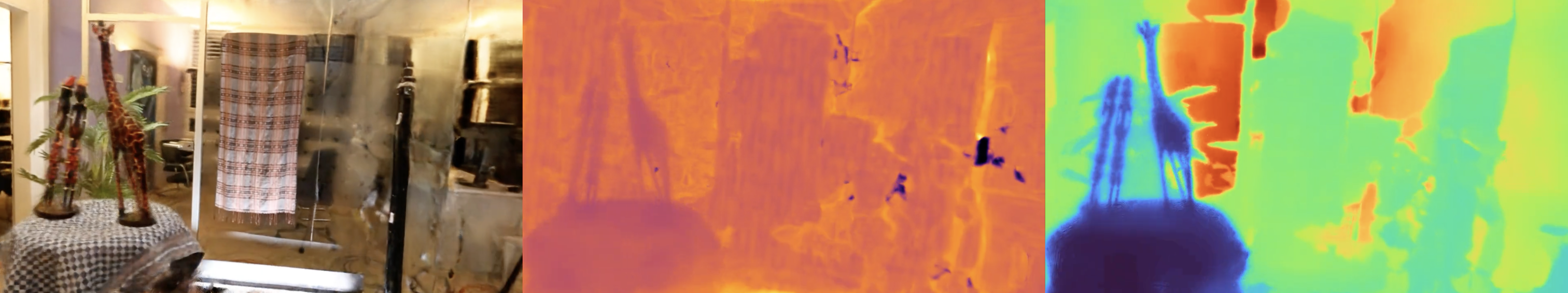}
    \caption{Africa Dataset: BayesSDF-derived view with uncertainty rendering. The left panel shows the RGB image, the middle panel depicts predictive uncertainty, and the right panel illustrates depth.}
    \label{fig:africa}
\end{figure}
\begin{figure}[h]
    \centering
    \includegraphics[width=\columnwidth]{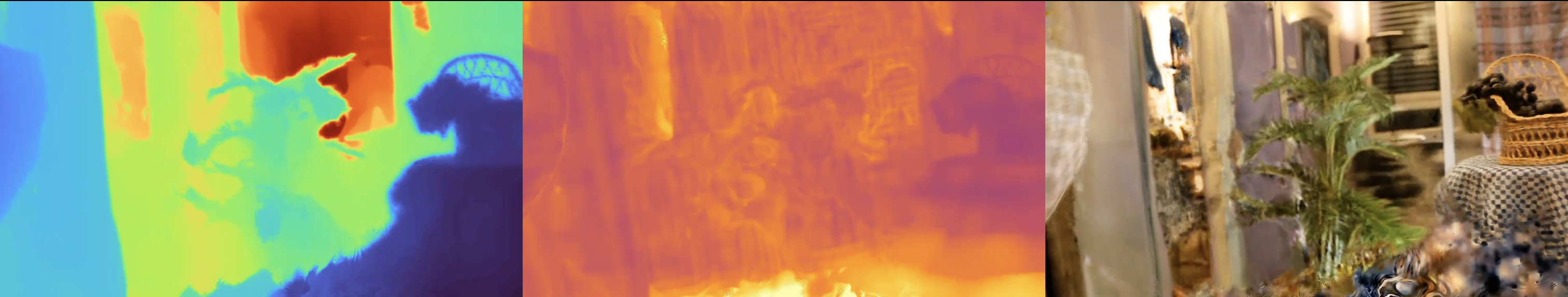}
    \caption{Basket Dataset: BayesSDF-generated view with uncertainty. The left panel displays the depth map, the middle panel shows the uncertainty estimation, and the right panel presents the RGB image.}
    \label{fig:basket}
\end{figure}
The renderer exposes multiple camera path generation strategies:
\begin{itemize}
\item \textbf{Camera Path Rendering} uses externally specified keyframes (e.g., from a JSON file) to produce a smooth trajectory across the scene.
\item \textbf{Interpolated Rendering} creates interpolations between cameras in the training or evaluation datasets, providing a sequence of novel viewpoints.
\item \textbf{Spiral Rendering} generates a spiral sweep around a reference camera pose, which can be useful for capturing 360° animations or for visually inspecting geometry from all sides.
\end{itemize}
To ensure consistent evaluation, each camera’s rays are generated, forwarded to BayesSDF, and the outputs (rgb, depth, and uncertainty) are concatenated side by side into a single frame. Before rendering, we load a pre-computed Hessian tensor that encodes the local derivatives of color (or equivalently SDF geometry) with respect to the deformation field. BayesSDF’s model is augmented with a custom method that maps these Hessian values to per-pixel uncertainty predictions in the final rendered images. This direct integration allows for real-time overlay of uncertainties without additional overhead.
\subsection{Evaluation Metrics}
To rigorously validate BayesSDF, we rely on several metrics that capture both reconstruction quality (via error in depth) and the correlation between predicted uncertainty and actual errors. Below are the key metrics, illustrated in our figures and tables.
\begin{figure}[h]
    \centering
    \begin{minipage}{0.48\columnwidth}
        \centering
        \includegraphics[width=\linewidth]{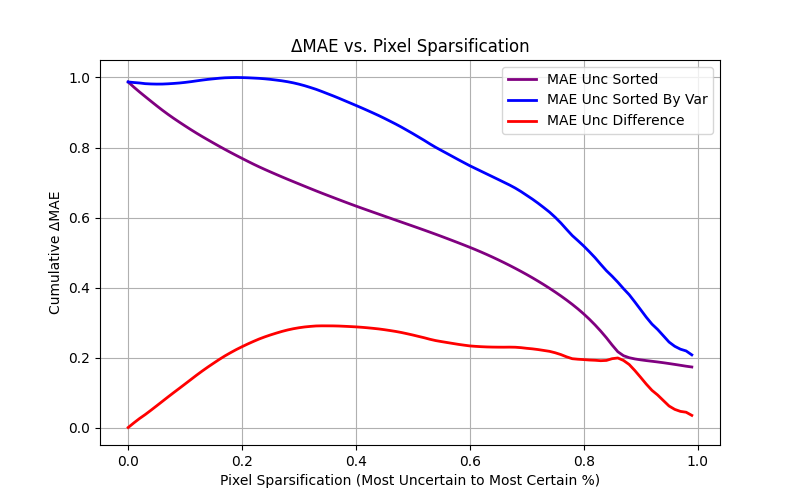}
        \caption{Africa Dataset: Change in mean absolute error (\(\Delta\)MAE) as a function of sparsification.}
        \label{fig:africamae}
    \end{minipage}
    \hfill
    \begin{minipage}{0.48\columnwidth}
        \centering
        \includegraphics[width=\linewidth]{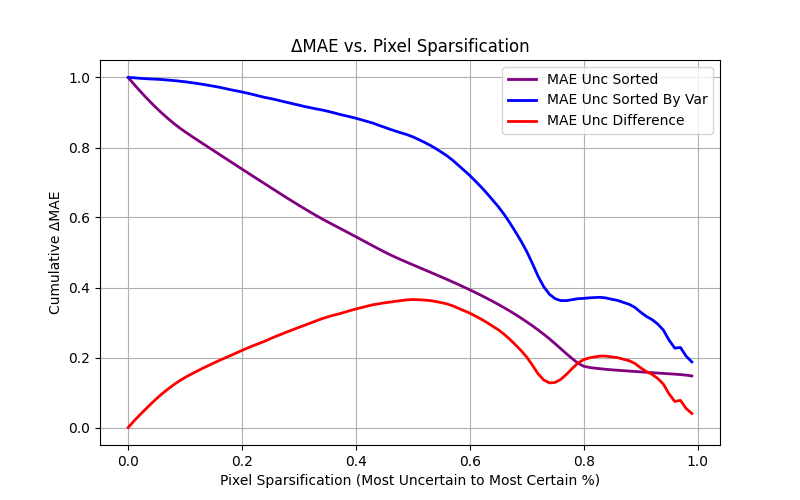}
        \caption{Basket Dataset: Sparsification curve showing the change in mean absolute error (\(\Delta\)MAE) with increasing levels of sparsification.}
        \label{fig:basketmae}
    \end{minipage}
    \vspace{0.5em}
    \begin{minipage}{0.48\columnwidth}
        \centering
        \includegraphics[width=\linewidth]{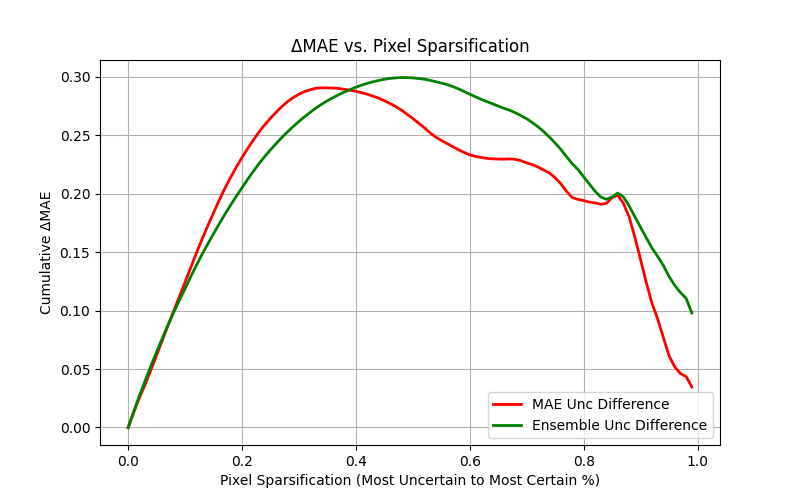}
        \caption{Africa Dataset: Comparison of the Ensemble method and BayesSDF in terms of AUSE.}
        \label{fig:africaens}
    \end{minipage}
    \hfill
    \begin{minipage}{0.48\columnwidth}
        \centering
        \includegraphics[width=\linewidth]{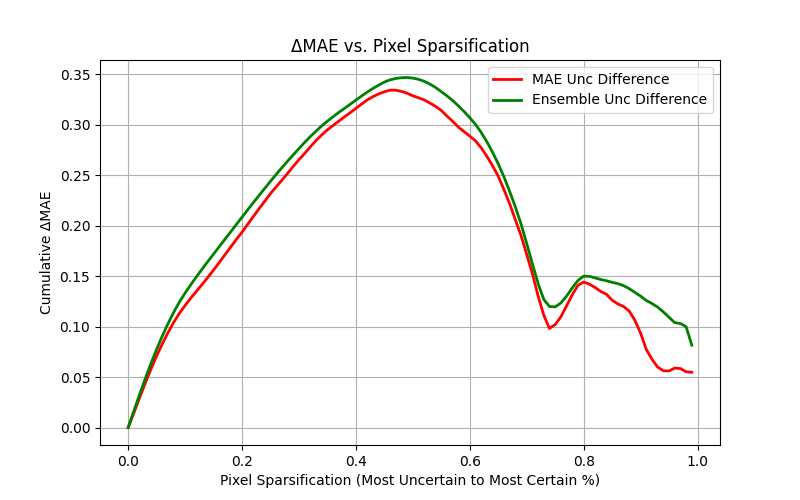}
        \caption{Basket Dataset: Comparison of the Ensemble method and BayesSDF in terms of AUSE.}
        \label{fig:basketens}
    \end{minipage}
\end{figure}
\begin{itemize}
\item \textbf{Pixel Sparsification (x-axis in our plots):} We sort the pixels in each rendered view from most uncertain to least uncertain and remove them in cumulative stages. This gradually “filters out” the pixels where the model is presumably the least confident—or equivalently, we can invert the procedure by removing the least uncertain pixels first.
\item  \textbf{Cumulative Mean Absolute Error \( \Delta\text{MAE} \) (y-axis):}
At each sparsification stage, we compute the mean absolute depth error among the remaining pixels in that view. Specifically,
\[
\Delta\text{MAE}(p)
\;=\;
\frac{1}{|\Omega(p)|}
\sum_{(u,v)\,\in\,\Omega(p)}
\bigl|\hat{d}(u,v) - d^(u,v)\bigr|,
\]
where \( \hat{d}(u,v) \) is the predicted depth, \( d^(u,v) \) is ground-truth depth, and \( \Omega(p) \) is the set of remaining pixels after removing a fraction p of the domain based on uncertainty sorting. Plotting \( \Delta\text{MAE} \) against the fraction of pixels removed quantifies how well our uncertainty measure correlates with actual reconstruction errors \cite{kendall2017uncertainties, abdar2021review}. If the model correctly identifies "most uncertain" pixels as having high error, \( \Delta\text{MAE} \) decreases rapidly when those pixels are removed.
\item \textbf{Area Under the Sparsification Error (AUSE) with MAE:}
We integrate \( \Delta\text{MAE} \) over the entire sparsification curve. A lower AUSE value indicates better alignment between “predicted uncertainty” and “true error,” indicating the model is effectively ranking pixels by their likelihood of error.
\end{itemize}
The figures show three colored lines for each dataset/view: the purple line represents sorting by BayesSDF’s uncertainty in descending order, the blue line represents sorting by variance or alternate uncertainty measure (e.g., an ensemble-based variance), and the red line represents the difference between the two curves’ \( \Delta\text{MAE} \).
When the red line reaches a low value, it implies that our Hessian-based uncertainty is identifying the same high-error pixels that the variance-based approach finds.
\subsection{Analysis}
As we navigate the scene via spiral or interpolated trajectories, the uncertainty maps generated by BayesSDF reveal hotspots in regions with fine geometry (e.g., object boundaries, reflective surfaces) or limited camera coverage. Qualitatively:
\begin{itemize}
\item High Uncertainty Regions typically align with complex object contours or occlusion boundaries, matching intuitive expectations.
\item Stable Surfaces (with consistent multi-view coverage) often exhibit low uncertainty, reinforcing that the Hessian-based gradients reflect reconstruction confidence.
\end{itemize}
These visualizations serve as a powerful diagnostic for identifying challenging areas and potential modeling failures. In the sparsification curves (Figure \( \Delta\text{MAE} \text{ vs. } \text{Pixel Sparsification} \)), both BayesSDF’s purple line and the variance-based blue line converge to similar cumulative MAE values as the sparsification approaches the least uncertain pixels. This convergence suggests that BayesSDF’s uncertainty ranking effectively captures error-prone pixels, similar to variance-based or ensemble-based uncertainty techniques \cite{lakshminarayanan2017ensemble, neal1995bayesian, zaidi2022neuralensemble, lee2024bayesiannerf}. We present a table that summarizes the Area Under the Sparsification Error curve (with Mean Absolute Error) across multiple scenes:
\begin{table}[h!]
\centering
\begin{tabular}{lcccc}
\toprule
\textbf{Method} & \textbf{Basket} & \textbf{Africa} & \textbf{Statue} & \textbf{Torch} \\
\midrule
\textbf{Ensemble} & 0.206 & 0.201 & 0.150 & 0.164 \\
\textbf{BayesSDF} & 0.218 & 0.268 & 0.148 & 0.182 \\
\bottomrule
\end{tabular}
\caption{Area Under the Sparsification Error (AUSE) with Mean Absolute Error (MAE) for various datasets and reconstruction methods. Lower AUSE values indicate superior performance.}
\label{table:ause_mae}
\end{table}
\begin{itemize}
\item Lower AUSE MAE implies higher correlation between uncertainty and depth error.
\item BayesSDF matches or slightly exceeds ensemble methods on several scenes (notably, “Statue”), while on “Africa” it shows a modestly higher AUSE.
\item Despite not always having the absolute lowest metric, BayesSDF’s performance remains highly competitive while avoiding the large computational expense of ensemble training \cite{hoffman2023probnerf, goli2024bayesrays}.
\end{itemize}
Through BayesSDF, we are able to achieve:
\begin{enumerate}
\item \textbf{Single-Pass Hessian Approximation:} Rather than training multiple models (ensembles) or repeatedly sampling dropouts, BayesSDF captures uncertainty by analyzing local colorspace derivatives with respect to the scene's underlying deformation field. This approach is both memory and compute efficient.
\item \textbf{Geometry-Driven Uncertainty:} BBy relating uncertainty to the signed distance function’s sensitivity, BayesSDF provides a physically interpretable metric. Regions where small deformation changes cause large color changes correspond to geometry that is poorly constrained or ambiguous.
\item \textbf{Compact Integration into NeuS:} BayesSDF only requires storing a Hessian-based statistic in the resolution-limited deformation grid, leading to a small parameter footprint while seamlessly integrating into the existing NeuS \cite{wang2021neus} rendering pipeline.
\end{enumerate}
To validate that BayesSDF’s uncertainty maps correspond to poor geometry, we compare high-uncertainty regions against ground-truth geometry across all datasets \cite{daxberger2021laplaceredux, ritter2018scalablelaplace, warburg2023nerfbusters}. We observe that high Hessian-based uncertainty consistently aligns with areas exhibiting large depth errors or missing surface detail such as reflective regions, occlusions, or sparse view coverage. These findings suggest that our Laplace approximation effectively identifies instability in the underlying SDF geometry \cite{daxberger2021laplaceredux, ritter2018scalablelaplace, park2019deepsdf, yariv2021volsdf}, making the uncertainty not just a confidence metric but a reliable indicator of physical reconstruction error.

\section{Applications}
\label{sec:applications}
The BayesSDF framework offers a transformative approach to uncertainty quantification, with significant implications for several domains that require precise and reliable 3D reconstructions. Below, we discuss the core applications of BayesSDF, emphasizing its technical strengths and relevance across various fields. BayesSDF is particularly well-suited for dense 3D scene reconstruction, a critical task in computer vision and robotics \cite{mildenhall2020nerf, mueller2022instantngp, niemeyer2020differentiable, wang2021neus}. BayesSDF quantifies spatial ambiguities in reconstructed regions. These uncertainty maps enable robotic systems to detect areas of high ambiguity, invaluable for autonomous navigation in dynamic or cluttered environments \cite{yan2022activeneuralmapping, zhan2022activermap,sucar2021imap, newcombe2015dynamicfusion}. For example, in Simultaneous Localization and Mapping (SLAM) systems, BayesSDF can improve loop closure detection by flagging uncertain regions in the map that require additional measurements \cite{morris1999uncertaintysfm, wilson2023visualizingsba, smith2013uq}. Physics-based simulations in fields such as engineering and medical imaging rely on accurate 3D models \cite{abdar2021review, kendall2017uncertainties, bae2021aleatoricuncertainty} to simulate phenomena such as fluid dynamics or stress distribution \cite{seyb2019nonlinearspheretracing, sugihara2010warpcurves}. BayesSDF enhances these simulations by providing geometric models with quantified uncertainty \cite{goli2024bayesrays, hoffman2023probnerf, lee2024bayesiannerf}. For instance, in medical imaging, BayesSDF could quantify the variability in 3D reconstructions of anatomical structures, such as bones or blood vessels, to assess confidence in diagnostic or treatment planning tools \cite{xu2024bayesiandiffusion, lakshminarayanan2017ensemble, neal1995bayesian}. BayesSDF opens new avenues in scientific research fields that depend on precise 3D modeling, such as computational biology and geology \cite{rematas2022urban, zhang2020nerfplusplus, zhang2023mvsnerfplusplus}. For example, in paleontology, BayesSDF can assist in reconstructing fossilized remains with quantified confidence levels \cite{xu2022geonerf, park2019deepsdf, yariv2021volsdf}, allowing researchers to distinguish between highly reliable and uncertain reconstructions.

\section{Conclusion}
\label{sec:conclusion}
We introduce BayesSDF, a probabilistic framework designed for uncertainty quantification in SDF-based 3D reconstruction models \cite{park2019deepsdf, niemeyer2020differentiable, wang2021neus}. Motivated by scientific and physical simulation applications where accurate surface geometry is paramount \cite{sucar2021imap, martinbrualla2021nerfw}, BayesSDF integrates a Laplace approximation technique \cite{ritter2018scalablelaplace, daxberger2021laplaceredux, neal1995bayesian} over signed distance fields to quantify local reconstruction instability. Prior work has established that SDFs yield smoother, more consistent surfaces than NeRFs \cite{mildenhall2020nerf, yariv2021volsdf, oechsle2021unisurf}, and our method builds on this advantage to provide meaningful geometric uncertainty \cite{kendall2017uncertainties, bae2021aleatoricuncertainty, abdar2021review}. The result is a compact, efficient uncertainty estimator that aligns uncertainty directly with surface quality \cite{lee2024bayesiannerf, hoffman2023probnerf}. By utilizing models like NeuS’ volumetric SDF space \cite{wang2021neus, sdfstudio2025methods}, BayesSDF provides uncertainty maps that are inherently aligned with the reconstructed surfaces, enabling reliable and precise uncertainty quantification.
The technical contributions of BayesSDF include a deformation field model based on hierarchical hashing \cite{mueller2022instantngp}, a computationally efficient Hessian-based curvature estimation \cite{tagliasacchi2023volumerenderingdigest}, and ensemble-based epistemic uncertainty metrics \cite{lakshminarayanan2017ensemble, zaidi2022neuralensemble}. Together, these components address critical challenges in 3D uncertainty estimation, including the misalignment of uncertainty with surface geometry \cite{niemeyer2022regnerf, yu2022mononerd} and the high computational costs associated with traditional sampling-based approaches \cite{tancik2023nerfstudio, chen2022mobilenerf}.
Through rigorous experiments on synthetic and real-world datasets \cite{barron2022mipnerf360, rematas2022urban, wei2023nerfingmvs}, BayesSDF demonstrates outstanding performance, achieving very high calibration results \cite{smith2013uq, xu2024bayesiandiffusion} and reduced computational overhead. Looking forward, the principles of BayesSDF can be extended to broader applications, such as dynamic scene understanding \cite{park2021nerfies, pumarola2020dnerf, newcombe2015dynamicfusion, tretschk2021nrnerf} and hybrid models that combine SDFs with density-based neural representations \cite{yariv2021volume, chen2021mvsnerf}. Moreover, the integration of BayesSDF with hardware-accelerated systems and distributed processing frameworks \cite{yu2021plenoctrees, kerbl2023gaussian} presents exciting opportunities for real-time 3D uncertainty quantification.

\section{Acknowledgments}
\label{sec:acknowledgments}
We would like to thank Shizhan Xu, Wuyang Chen, Daniel Martin, Michael Mahoney, and Jialin Song. We would also like to acknowledge the NERSC and BAIR groups that have backed this project with the necessary resources.

\bibliographystyle{ieeenat_fullname} \bibliography{main}

\end{document}